\documentclass{article}

\usepackage[dvipsnames]{xcolor}
\usepackage{spconf,amsmath,graphicx}
\usepackage{adjustbox}
\usepackage{multirow}
\usepackage{float}
\usepackage{booktabs}
\usepackage{array}
\usepackage{url}
\usepackage{algorithm}
\usepackage{algpseudocode}
\usepackage{enumitem}
\usepackage{cite}
\usepackage{hyperref}
\usepackage{cleveref}
\usepackage{multirow}
\usepackage{comment}
\usepackage{makecell}
\usepackage{lipsum}
\usepackage[normalem]{ulem}

\definecolor{mygreendark}{HTML}{82B366}
\definecolor{myorangedark}{HTML}{D79B00}
\definecolor{mybluedark}{HTML}{6C8EBF}
\definecolor{myyellowdark}{HTML}{D6B656}
\definecolor{myreddark}{HTML}{B85450}
\definecolor{mypurpledark}{HTML}{9673A6}

\newcolumntype{+}{>{\global\let\currentrowstyle\relax}}
\newcolumntype{^}{>{\currentrowstyle}}
\newcommand{\rowstyle}[1]{\gdef\currentrowstyle{#1}%
  #1\ignorespaces
}

\newcommand\extrafootertext[1]{%
    \bgroup
    \renewcommand\thefootnote{\fnsymbol{footnote}}%
    \renewcommand\thempfootnote{\fnsymbol{mpfootnote}}%
    \footnotetext[0]{#1}%
    \egroup
}

\title{Are GAN-based Morphs Threatening Face Recognition?}
\name{Eklavya Sarkar$^{1,2}$, Pavel Korshunov$^{1}$, Laurent Colbois $^{1,3}$, and S\'ebastien Marcel$^{1,3}$}
\address{$^{1}$Idiap Research Institute, Martigny, Switzerland\\
$^{2}$École polytechnique fédérale de Lausanne, Switzerland\\
$^{3}$University of Lausanne, Switzerland\\
{\tt\small \{eklavya.sarkar, pavel.korshunov, laurent.colbois, sebastien.marcel\}@idiap.ch}
}
\graphicspath{{figures/}}
\begin{document}
%\ninept
%
\maketitle
%
%%%%%%%%% ABSTRACT
\begin{abstract}
    Morphing attacks are a threat to biometric systems where the biometric reference in an identity document can be altered. This form of attack presents an important issue in applications relying on identity documents such as border security or access control. Research in generation of face morphs and their detection is developing rapidly, however very few datasets with morphing attacks and open-source detection toolkits are publicly available. This paper bridges this gap by providing two datasets and the corresponding code for four types of morphing attacks: two that rely on facial landmarks based on OpenCV and FaceMorpher, and two that use StyleGAN 2  to generate synthetic morphs. We also conduct extensive experiments to assess the vulnerability of four state-of-the-art face recognition systems, including FaceNet, VGG-Face, ArcFace, and ISV. Surprisingly, the experiments demonstrate that, although visually more appealing, morphs based on StyleGAN 2 do not pose a significant threat to the state to face recognition systems, as these morphs were outmatched by the simple morphs that are based facial landmarks.
    %\textcolor{green}{The experiments demonstrate that GAN-based morphs generated with a sophisticated loss pose considerably more of a threat to face recognition systems compared to na\"ive linear-interpolation based ones, but nonetheless significantly less than traditional landmark-based morphs}.
\end{abstract}

%%%%%%%%% BODY TEXT
%
\section{Introduction}
\label{sec:introduction}

\extrafootertext{Work funded by the Swiss Center for Biometrics Research and Testing.}

\begin{figure*}[ht]
\begin{minipage}[b]{0.16\linewidth}
  \centering
  \centerline{\includegraphics[width=\textwidth]{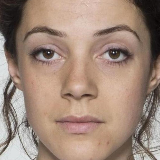}}
  \centerline{(a) Identity A}\medskip
\end{minipage}
\begin{minipage}[b]{0.16\linewidth}
  \centering
  \centerline{\includegraphics[width=\textwidth]{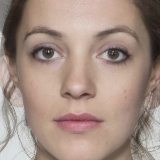}}
  \centerline{(b) OpenCV}\medskip
\end{minipage}
\begin{minipage}[b]{0.16\linewidth}
  \centering
  \centerline{\includegraphics[width=\textwidth]{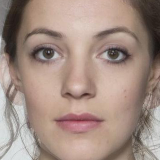}}
  \centerline{(c) FaceMorpher}\medskip
\end{minipage}
\begin{minipage}[b]{0.16\linewidth}
  \centering
  \centerline{\includegraphics[width=\textwidth]{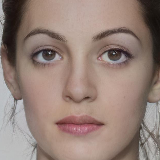}}
  \centerline{(d) StyleGAN2}\medskip
\end{minipage}
\begin{minipage}[b]{0.16\linewidth}
  \centering
  \centerline{\includegraphics[width=\textwidth]{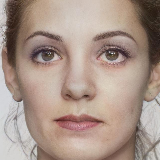}}
  \centerline{(e) MIPGAN-II}\medskip
\end{minipage}
% \begin{minipage}[b]{0.12\linewidth}
%   \centering
%   \centerline{\includegraphics[width=\textwidth]{facelab_london/preprocessed/webmorph_001_002.png}}
%   \centerline{(f) WebMorph}\medskip
% \end{minipage}
% \begin{minipage}[b]{0.12\linewidth}
%   \centering
%   \centerline{\includegraphics[width=\textwidth]{facelab_london/preprocessed/amsl_001_002.png}}
%   \centerline{(g) CombinedMorph}\medskip
% \end{minipage}
\begin{minipage}[b]{0.16\linewidth}
  \centering
  \centerline{\includegraphics[width=\textwidth]{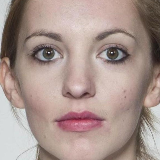}}
  \centerline{(h) Identity B}\medskip
\end{minipage}
\caption{Different types of generated morphed images from two identities in the FRLL dataset.}
\label{fig:morph_figures}
\end{figure*}

% \begin{figure*}[ht]
% \begin{minipage}[b]{\linewidth}
%   \centering
%   \centerline{\includegraphics[width=\textwidth]{Block/ICASSP_Block_BW.pdf}}
% \end{minipage}
% \caption{Morphed image generation pipeline for landmark-based methods.}
% \label{fig:block}
% \end{figure*}

After Ferrara \emph{et al.}~\cite{Ferrara2014} showed that by using a morphed photo of two different people an adversary can circumvent passport registration process, morphing attacks and how to detect them received a lot of attention from academic, industrial, and security communities. The vulnerability of state-of-the-art (SOTA) face recognition systems (FR) and the threat such vulnerability poses to the security systems relying on recognition technologies led to the explosion of research work in this area.

Most of the work related to morphing attacks (MAs) focuses on their detection. Recently proposed techniques for morphing attack detection (MAD) include methods based on \emph{so called} classical approaches using local binary patterns (LBP) and support vector machines (SVM)~\cite{Spreeuwers2018}, approaches rooted in image forensics that rely on photo response non uniformity (PRNU) function~\cite{Scherhag2019prnu}, deep neural networks specifically trained to detect morph images~\cite{Seibold2020}, and FR systems themselves serving as feature extractors for an support vector machine (SVM) classifier~\cite{Wandzik2018}. The National Institute of Standards and Technology (NIST) is now conducting independent evaluations of MAD technologies~\cite{NIST2020}.

However the research in the area of morphing attacks and their detection suffers from a lack of 
%standardisation of morph generation tools, 
datasets, evaluation protocols, and clear understanding of whether the latest face recognition systems are vulnerable to both `classical' and the latest generative adversarial network (GAN)-based morphing attacks. 
% We identify two main issues:
So called `classical' landmark-based morphing techniques are widely available, but the modern ones are rarely publicly released. Novel methods are often either proprietary, such as Combined Morphs \cite{StirTrace}, or are difficult to 
%else are only described on paper. Those which use deep learning frameworks based on a graph of operators such as TensorFlow, which are often used in the latest GAN-based morphs, are complex to decode and successfully 
replicate from a published description without knowing the minutes technical details. Pre-generated databases of morphing attacks are therefore essential for biometrics research, yet only a few, like the Face Morph Image dataset~\cite{StirTrace} by Advanced Multimedia Security Lab, are publicly available. 

% Even if open-sourced and implemented with ease-of-access, modern SOTA GAN-based models typically require high-end computational power and memory.

% iii) The protocols which define which specific images to enroll as reference, and then compare to other specific probes are often are quite limited.

% As a consequence of this lack of shared resources, most papers implement their own systems and databases, making it very hard to consistently compare results of vulnerability analysis of morphing attack detection across papers. 

Therefore, this paper provides the following contributions:

\begin{enumerate}
    \item We provide an open source morphing tool\footnote{\href{https://gitlab.idiap.ch/bob/bob.paper.icassp2022\_morph\_generate}{https://gitlab.idiap.ch/bob/bob.paper.icassp2022\_morph\_generate}} for generation of morphing attacks based on OpenCV~\cite{opencv}, FaceMorpher~\cite{facemorpher_repo}, StyleGAN 2~\cite{stylegan2}, and our modified implementation of MIPGAN-II~\cite{MIPGAN2021}.
    \item We provide publicly available datasets\footnotemark[1] with morphed images generated using the aforementioned techniques, including the latest GAN-based morphs, on the publicly available FERET~\cite{feret} and Face Research Lab London (FRLL)~\cite{london} datasets.
    \item We conduct an extensive vulnerability assessment of the different morphing attacks images in our generated dataset against SOTA face recognition systems, used in some of the latest morphing attacks vulnerability studies~\cite{deep_face_representations}. We specifically use the FaceNet~\cite{facenet} VGG-Face~\cite{deepfacerecognition}, ArcFace~\cite{arcface_paper}, and ISV~\cite{isv} systems, which are pre-trained on `clean' bona fide databases. 
\end{enumerate}

We also highlight that the majority of existing work on Morphing Attacks (MAs) only consider the `typical' scenario where the morphs are used to attack the enrollment process of face recognition. This paper, to the best of our knowledge is amount the the first to evaluate two scenarios: i) when morphs attack the the enrollment process, and ii) when they are used to attack the probing process, which is similar to a presentation attack~\cite{iso_iampr}. 

%We also compare our results with other papers in the literature and provide a discussion on the similar trends and differences between the scores.

% We strongly believe the contributions of this paper prove to be useful to the biometrics community, and will significantly help verify, reproduce, extend, and accelerate the work of all researchers in this field.

\section{Morph Generation}
\label{sec:morph_generation}

In this section, we present the datasets with bona fide faces and the different tools, including GAN-based, that we use to generate the morphing images for the vulnerability experiments.

\subsection{Datasets}
\label{ssec:database}
We used FERET~\cite{feret} and FRLL~\cite{london} datasets of facial images to generate the morphs. FERET was selected because it is the \emph{de facto} the standard datasets commonly used in papers on morphing attack detection~\cite{pad,deep_face_representations} and it has large number of images of different identities. The FRLL dataset is also ideal for creating morphing attacks because it contains close-up frontal face images of very high visual quality and $1350 \times 1350$ resolution, shot under \textit{uniform} illumination with large varieties in ethnicity, pose, and expression. Each face is annotated using $189$ facial landmarks, which is notably a very high number, as typical landmarks detectors provide no more than $68$-$70$ landmarks. The main limitation of FRLL dataset, compared to FERET, is the limited number $102$ of different identities with $53$ males and $49$ females.

For each dataset, we select bona fide (or original) face pairs for morph generation by following the existing protocols used in previous work. For FERET, we follow the protocols used in the work by Scherhag \emph{et al.}~\cite{deep_face_representations} that were kindly provided by the authors.
%\textcolor{red}{(\sout{though they were not able to provide any morphed images)}}. 
For FRLL dataset, we follow the protocols used in AMSL Face Morph Image dataset by Neubert \emph{et al.}~\cite{StirTrace}. Using these protocols (essentially, which facial image pairs to morph), we generated morphs using four different methods: based on OpenCV, based on FaceMorpher, based on StyleGAN 2, and a modified MIPGAN-II~\cite{MIPGAN2021}.
%, which are described in Section~\ref{ssec:morphing_tools}. 

\subsection{Morphing Tools}
\label{ssec:morphing_tools}
As representatives of the `classical` landmark-based morphing tools, we provide two commonly used open source face morphing algorithms. First one is the \textbf{OpenCV}-based algorithm, referred throughout the paper as \textit{OpenCV}, which is an adaptation of an open-source implementation~\cite{opencv} for morphing faces using $68$-point annotator from Dlib library~\cite{dlib09}. Face landmarks are obtained for each of the two bona fide source images and are used to form Delaunay triangles, which are in-turn warped and alpha blended.

\textbf{FaceMorpher}~\cite{facemorpher_repo}, referred to as \textit{FaceMorpher}, is another open-source landmark-based morphing algorithm, but with the STASM~\cite{stasm} landmark detector instead. Both algorithms create morphs with noticeable ghosting artefacts for all three datasets, as the region outside the area covered by these landmarks is simply averaged.

% \subsubsection{Generative Adversarial Network-Based Morphs}
% \label{sssec:gan_based_morphs}
Following the advances in generative adversarial networks (GANs), there were attempts to generate morphed images using a GAN instead of landmark-based methods~\cite{GAN_Morphs_Threat,korshunov2019deepmorphs}. In this paper, we adapted the latest~\textbf{StyleGAN 2}~\cite{stylegan2}, referred throughout the paper as \textit{StyleGAN2}, to develop a morphing algorithm which can generate high resolution realistic looking faces with no noticeable artifacts. The StyleGAN 2 was pre-trained on the FFHQ dataset introduced in~\cite{stylegan}. 

The faces are cropped to obtain the same landmark alignment as in the FFHQ dataset. The images are then projected into the $\mathcal{W}$ space of StyleGAN 2 by optimizing the input latent style vector that is fed to the generator network, such that it minimizes the perceptual loss between the generated and real image~\cite{stylegan2}. Once an associated latent vector has been computed for each of the source images, morphs can be generated by linearly interpolating between two latent vectors, and feeding the interpolated vector back into the generator.

This technique yields very realistic looking morphs without visual artefacts, however, since StyleGAN does not have any information about the identities in bona fide images, there is no guarantee that the resulted morph is actually a blend of these identities (see the example in Figure~\ref{fig:morph_figures}(d) for an idea). 

For this purpose, \textbf{we implemented a modified version} of the recent MIPGAN-II technique~\cite{MIPGAN2021}, referred throughout the paper as \textit{MIPGAN-II}, which improves on the StyleGAN 2 morphs, by \textit{further} optimizing the interpolated latent vector with four additional weighted losses, which help to preserve the identity information and structural correspondence of the two bona-fide images. The main difference in our versions of MIPGAN is that we use the pre-trained VGGFace model with the ResNet50 backbone as feature extractor in the identity loss, instead of a pre-trained embedding extractor with ResNet50 as backbone using the ArcFace loss.

Both, StyleGAN2 and MIPGAN-II GAN-based morphs require the projected images to be at a high resolution ($1024 \times 1024$ after cropping), and work better with an uniform background, which makes the FRLL dataset particularly appropriate. A side note observation of using GAN-based techniques for generating morphs is that it is equally easy to generate high-quality morphs for smiling expressions as it is for the neutral faces, which is not possible with typical landmark-based tools.

Using four (two `classical' and two GAN-based) morphing tools, we generate $529$ morphs per each tool for FERET datasets using the same protocols as in~\cite{deep_face_representations}, and $1222$ morphs per tool images for FRLL, following the morph generation protocol defined in ASML Face Morph Image dataset. It is to be noted that the protocols insure not to morph across genders and ethnicities, and only interpolate images if neither or only one of the two subjects is wearing glasses.

% We publicly release all four of the described methods as our open-source toolkit for morph generation as one of the three main contributions of this paper. We additionally implement the ability to generate morphs with different ratios of the two bona fide identities.

\section{Evaluation Protocols}
\label{sec:evaluation_protocol}

\subsection{Face Recognition Systems}
\label{ssec:frs_pipelines}
To evaluate vulnerability of face recognition against morphing attacks, we used publicly available pre-trained FaceNet \cite{facenet}, ArcFace~\cite{arcface_paper}, and VGG-Face~\cite{deepfacerecognition}  architectures. We used the last fully connected layers of these networks as features and the cosine distance as a classifier. For a given test face, the confidence score of whether it belongs to a reference model is the cosine distance between the average reference feature vector and the feature vector of a test face. These systems are the state of the art recognition systems with Facenet showing $99.63\%$~\cite{facenet}, ArcFace -- $99.53$~\cite{arcface_paper}, and VGG-Face -- $98.95\%$~\cite{deepfacerecognition} accuracies on the labeled faces in the wild (LFW) dataset.

We also used an inter-session variability (ISV) based face recognition~\cite{isv}, pre-trained on the MOBIO~\cite{mobio} dataset, as a `classical' baseline. The DCT features computed on overlapping blocks of $40 \times 40$ were used for the ISV-based system of $512$ Gaussian mixture models (GMMs) and $160$ dimensional subspace.

\subsection{Evaluation Metrics}
\label{ssec:evaluation_metrics}
In a verification process, the user attempting to authenticate presents a biometric probe and a claimed identity, and can be classified into one of the following 3 categories. \textit{A) Genuine user} (BF): probe and claimed identity both correctly belong to the user. \textit{B) Zero-effort impostor} (BF): probe belongs to the user, but the claimed identity corresponds to a different enrolled user. \textit{C) Morph attack impostor} (MA): probe matches the claimed identity but does not correspond to the user.

The \textit{verification} performance is typically evaluated with the following metrics. 
\begin{itemize}
    \item \textit{False Match Rate (FMR)}~\cite{pad}: proportion of zero-effort impostors that are falsely authenticated.
    \item \textit{False Non-Match Rate (FNMR)}~\cite{pad}: proportion of genuine users which are falsely rejected.
    \item \textit{Mated Morph Presentation Match Rate  (MMPMR)}~\cite{Biometric_Systems_under_Morphing_Attacks}: proportion of morphs attacks impostors accepted by the face recognition system.
\end{itemize}

\subsection{Evaluation scenarios}
\label{ssec:scenarios}
% We adopted the same evaluation scenarios used in~\cite{deep_face_representations} for the FERET dataset, but defined our own for FRLL due to the lack of publicly available protocols for this dataset.

In general, there are two main scenarios under which a face recognition system is evaluated: a bona fide (BF) scenario where both the reference and probes images as genuine, so there are no attacks and the system is assumed to perform under the conditions it was designed for; and the morphing attack (MA) scenario when morphs are introduced to the face recognition with a malicious intent to spoof the recognition. There are also two variants of MA scenario, when a morphed image can be either used as a reference, i.e., FR system is hijacked during enrollment process (typical morphing attack scenario), or a morphed image is used as a probe, which is similar to presentation attack scenario~\cite{iso_iampr}. 

The number of reference and probe images for each evaluation scenario is summarized in Table~\ref{table:comparaisons}.

\begin{table}[ht]
    \centering
    \caption{Number of images in different evaluation scenarios.}
    \label{table:comparaisons}
    % \begin{adjustbox}{width=\columnwidth,center}
    \begin{tabular}{+l^l^l^l^l^l}
    \toprule
    \rowstyle{\bfseries} Dataset & Morphs as & BF & MA & Impostors \\
    \midrule
    \multirow{2}{*}{FERET} & References & 529 & 791 & 418,439 \\ % for_probes.lst
                          & Probes     & 791 & 529 & 418,439 \\ % for_probes.lst
    \midrule
    \multirow{2}{*}{FRLL}  & References & 91  & 584 & 1,984 \\ % because we use for_scores.lst (fewer comparaisons)
                          & Probes     & 584 & 91  & 4,153 \\ % for_probes.lst (more comparaisons) -- I get these numbers from the length of the scores files
    % \midrule
    % \multirow{2}{*}{FRGC} & References  & 3,298 & 964 & 1,698,384 \\ % for_probes.lst
    %                       & Probes      & 964 & 3,298 & 1,698,384 \\ % for_probes.lst
    \bottomrule
    \end{tabular}
    % \end{adjustbox}
\end{table}

It is also to be noted that we did not split datasets into training, development, and test subsets but used each whole dataset as one single test set, as all used FR systems were pre-trained on other databases. Furthermore, we choose the decision threshold to compute MMPMR value for MA scenario based on FMR value computed in the bona fide scenario, thus removing the need for a development set.

\section{Experimental Results}
\label{sec:experimental_results}

Table~\ref{table:all_iamprs} summarizes the results of the vulnerability assessment of the several face recognition systems (described in section~\ref{ssec:frs_pipelines}) under the different morphing attack scenarios (as explained in section~\ref{ssec:scenarios}). The MMPMR metric is calculated by setting the decision threshold at FMR=$0.1$\% in the bona fide scenario.

% - Thresholds, FAR criteria, licit-dev
% - Trends
% - Morphs as references or probes roughly give the same IAMPR (except gabor where morphs as probes is significantly more vulnerable)
% - FFLR seems most vulnerable, then FERET, then FRGC
% - OpenCV, then FaceMorpher, then StyleGAN most vulnerable
% - OpenCV and Facemorpher are roughly the same everywhere, except for Gabor
% - StyleGAN 2 morphs as significantly less vulnerable than the rest (at 0.1% FAR, but what about for 1% FAR) 
% --- Message: StyleGAN2-morphs don't present a sign of vulnerability on these 3 datasets. Basic linear interpolation is not enough to fool FRS.

\begin{table}[ht]
    % \centering
    % \captionsetup{justification=centering}
    \caption{MMPMR @ FMR = 0.1\% \newline (Morphs as references | Morphs as probes) [\%]}
    % \begin{adjustbox}{width=\columnwidth
    \label{table:all_iamprs}
    \begin{tabular}{+l^l^c^c}
    \toprule
    Tools & FRS & FRLL & FERET \\ 
    \midrule
    \multirow{3}{*}{OpenCV} & FaceNet & 83.3 | 72.0 & 41.1 | 40.6 \\
                            & Arcface & 59.8 | 73.8 & 34.6 | 35.2 \\
                            & VGG     & 39.7 | 48.6 & 22.0 | 21.0 \\
                            & ISV     & 59.8 | 97.8 & 44.8 | 58.4 \\
    \midrule
    \multirow{4}{*}{FaceMorpher} & FaceNet & 64.5 | 68.2 &  39.9 | 40.3 \\
                                 & Arcface & 57.6 | 75.3 &  34.1 | 34.8 \\
                                 & VGG     & 23.4 | 47.1 &  20.5 | 18.3 \\
                                 & ISV     & 56.1 | 96.1 &  42.6 | 56.5 \\
                                 
    \midrule
    \multirow{4}{*}{StyleGAN2} & FaceNet & 5.9  | 11.0 & 1.6 | 1.3 \\
                               & Arcface & 9.8  | 18.3 & 2.4 | 2.5 \\
                               & VGG     & 3.0  | 9.1  & 2.0 | 1.5 \\
                               & ISV     & 9.2  | 43.6 & 2.7 | 3.4 \\
    \midrule
    \multirow{4}{*}{MIPGAN-II} & FaceNet & 47.2 | 62.7 & 32.9 | 32.3 \\
                               & Arcface & 32.0 | 46.5 & 26.0 | 25.1 \\
                               & VGG     & 15.9 | 30.4 & 14.5 | 13.2 \\
                               & ISV     & 3.6  | 23.7 & 7.3  | 9.6 \\
    \bottomrule
    \end{tabular}
    % \end{adjustbox}
\end{table}

The results in Table~\ref{table:all_iamprs} reveal a number of interesting observations. The StyleGAN2-morphs do not pose a significant threat to the state of the art face recognition systems, compared to landmark-based morphs, despite being of higher visual quality, and with very few ghosting artefacts. This likely occurs because the original pixels of both contributing images are still present in the features after landmark-based morph-generation pipeline is applied, and are later picked up during face recognition, thus successfully fooling the FR systems. Conversely, the StyleGAN pipeline conserves no pixel traces of the original contributing subjects, other than the positions of the facial landmarks, as it generates the morphed image by interpolating the projected vectors in the $\mathcal{W}$ latent space. The interpolated vector fed back through the synthesis network does not contain the features of both identities, and instead is perceived as a new, different identity altogether.

This is further proved when the MIPGAN-II morphs which purposefully use four additional losses to further optimize the generated morph in an attempt to conserve the identities of the two source subjects: the vulnerability is significantly higher than with naive linear-interpolation in the StyleGAN $\mathcal{W}$ space. However, our implementation of MIPGAN-II lead to twice as low MMPMR rates compared to the numbers reported in original MIPGAN-II paper~\cite{MIPGAN2021}. Such significant disparity in results is puzzling but a few elements could contribute to it:

A) It appears that rather than using the $\mathcal{W}$ space of StyleGAN 2 for generating the morphs, as we did in our implementation, the authors of MIPGAN-II paper~\cite{MIPGAN2021} instead used the $\mathcal{W+}$~\cite{9008515} latent space of StyleGAN, as they seem to describe their latent projections as a concatenation of $18$ different $512$-dimensional $w$ vectors, one for each layer of the StyleGAN architecture. We believe that using $\mathcal{W}$ space is more reasonable as operating in $\mathcal{W+}$ does not guarantee a visual realism of resulted morphs. However, this point is hard to verify, since the authors of MIPGAN-II~\cite{MIPGAN2021} did not release their code and their paper is not very explicit about this aspect.
    % Since the authors do not release their code and do not specify in their paper, we can only speculate that they interpolated in the higher-dimension, extended $\mathcal{W+}$ latent space of StyleGAN, which is a concatenation of 18 different 512-dimensional $w$ vectors, one for each layer of the StyleGAN architecture rather than the $\mathcal{W}$ space as we did.

B) The pre-trained StyleGAN model in~\cite{MIPGAN2021} was fine-tuned on their test dataset (FRGCv2 \cite{frgc}), which made the generated morphs to appear visually and structurally very similar to the original images in the dataset. This type of `trick' clearly would increase the chanced of the morphs to be more threatening to face recognition, which was tested on the same dataset.
    %visual/geometrical/structural resemblance to the contributing dataset, which then in turn significantly increased the vulnerability scores to similar levels as the landmark based morphs. 

% C) The identity loss component of MIPGAN-II~\cite{MIPGAN2021} used ArcFace network with ResNet50 architecture to make the generated morphs appear more similar to the target and source faces. However, the same network was used as one of the face recognition systems, for which MMPMR values were reported as especially high. Using the same model for both generation and detection creates a certain bias that we believe led to such high vulnerability rates. In our implementation of MIPGAN-II, we used a VGGFace model instead and VGG-16 model (is it different??) in the face recognition experiments.
    %pre-trained with triplet-loss and a ResNet50 architecture is used for the image feature extractor in the identity loss instead of a ResNet50 pre-trained using the ArcFace loss. %{and the VGG-16 model is used as extractor for the vulnerability assessment instead of the same ResNet50.

Table~\ref{table:all_iamprs} also demonstrates that the more accurate face recognition system is the more vulnerable it is to morphing attacks, which is also in line with the findings reported for presentation attacks~\cite{deeply_vulnerable}. This trend is especially evident when we compare a more accurate and deeper FaceNet architecture with VGG for all databases and types of morphs.

We can also notice that the results for the scenario when morphs attack the enrollment process (see `morphs as references' sub-columns in Table~\ref{table:all_iamprs}) are very similar to the results for the scenario when morphs are used as probes (see `morphs as probes' sub-columns) for the FERET morphs database. However, in the case of FRLL, the face recognition systems are clearly more vulnerable to the scenario when morphs are used as probes. It means that the quality of original images used to create morphs may lead to more threatening morphs in the presentation attack scenarios, rather than when attacking FR systems from the inside. 

%We also know that the protocols used to generate the FRLL morphs specifically did not mix across gender and ethnicities, suggesting that a pre-selection of the contributing subjects can indeed improve performance against FR systems.

\section{Conclusion}
\label{sec:conclusion}
In this paper, we assess the level of vulnerability of existing face recognition systems, based on VGG-Face, ArcFace, and FaceNet neural network models, against four morphing attacks, including two `classical' morphs based on facial landmarks and two based on StyleGAN 2. The results demonstrate that `classical' morphs still are of the highest threat to the face recognition while GAN-based morphs, despite their higher visual appeal, do not pose as much of a thread to automated system. We also note that the face recognition systems that are better at recognition are also more vulnerable to morphing attacks. We publicly release the databases we generated and used, and provide all tools for generating morphs and running the evaluation experiments as an open source package.

%We compare our results with others in the literature and provide a discussion. provides three major contributions to the morphing attacks field. We release an open-source morph generation toolkit containing the OpenCV, FaceMorpher, StyleGAN, and modified MIPGAN-II techniques for the scientific community in order to streamline and standardise research using these methods. We also publicly release the databases with pre-generated morphs using said methods as currently very few public datasets of morphs exist. Finally, we assess the level of vulnerability of existing FR systems, based on VGG-Face, ArcFace, and FaceNet neural network models, against various morphing attacks coming from datasets completely distinct from those used to train the FR systems. We compare our results with others in the literature and provide a discussion. The experiments demonstrate that [...].

% \section{Acknowledgments}
% \label{sec:acknowledgments}
% We thank the NCCR Evolving language, funded by the Swiss National Science Foundation (SNSF) through the grant 51NF40 180888, the Swiss Center for Biometrics for Research and Testing, and Idiap Research Institute for supporting this research.

%\vfill
\clearpage
\bibliographystyle{IEEEbib}
% \bibliography{refs}
{\footnotesize \bibliography{refs}} % Uncomment to make smaller

\end{document}